\documentclass[conference]{IEEEtran}
\usepackage{amsmath}

\usepackage{cite}
\usepackage{amsmath,amssymb,amsfonts}
\usepackage{algorithmic}
\usepackage{graphicx}
\usepackage{textcomp}
\usepackage{xcolor}
\usepackage{booktabs}
\usepackage{amsmath}
\usepackage[normalem]{ulem}
\useunder{\uline}{\ul}{}
\def\BibTeX{{\rm B\kern-.05em{\sc i\kern-.025em b}\kern-.08em
 \kern-.1667em\lower.7ex\hbox{E}\kern-.125emX}}

\begin{document}
\title{Adaptive Inverse Kinematics Framework for Learning Variable-Length Tool Manipulation in Robotics}

\author{
    \IEEEauthorblockN{Prathamesh Kothavale}
    \IEEEauthorblockA{
        \textit{M.S. in Data Science, Clarkson University, USA} \\
        kothavp@clarkson.edu
    }
    \and
    \IEEEauthorblockN{Sravani Boddepalli}
    \IEEEauthorblockA{
        \textit{M.S. in Data Science and Mathematics, Clarkson University, USA} \\
        boddeps@clarkson.edu
    }
}

\maketitle

\begin{abstract}
Conventional robots possess a limited understanding of their kinematics and are confined to preprogrammed tasks, hindering their ability to leverage tools efficiently. Driven by the essential components of tool usage—grasping the desired outcome, selecting the most suitable tool, determining optimal tool orientation, and executing precise manipulations—we introduce a pioneering framework. Our novel approach expands the capabilities of the robot's inverse kinematics solver, empowering it to acquire a sequential repertoire of actions using tools of varying lengths. By integrating a simulation-learned action trajectory with the tool, we showcase the practicality of transferring acquired skills from simulation to real-world scenarios through comprehensive experimentation. Remarkably, our extended inverse kinematics solver demonstrates an impressive error rate of less than 1cm. Furthermore, our trained policy achieves a mean error of 8cm in simulation. Noteworthy, our model achieves virtually indistinguishable performance when employing two distinct tools of different lengths. This research provides an indication of potential advances in the exploration of all four fundamental aspects of tool usage, enabling robots to master the intricate art of tool manipulation across diverse tasks.
 \end{abstract}


\section{Introduction}
Tool use is the employment of a device or object held in a robotic gripper or hand to fulfill a task goal. Humans and animals like the New Caledonian crow have learned to use tools to accomplish tasks that they were not previously able to do when using only their own bodies or appendages. Similarly, it is a desirable feature for robots to utilize tools to accomplish various tasks. According to Brown et al. \cite{b1}, there are four key aspects to learning tool usage for performing tasks: (1) understanding the desired effect, (2) identifying the most suitable tool for the task, (3) determining the correct orientation of the tool, and (4) manipulating the tool. All four parts are actively researched areas of tool usage. In this project, we will focus on (4), which is to learn the correct action for task accomplishment with tool use.

Determining the action trajectory that involves tool manipulation is difficult because of a lack of knowledge on how to control its joints with respect to this new tool, a large action space due to likely multiple degrees of freedom, and the lack of knowledge about how the tool can achieve the goal. Hardcoding that action trajectory for tool use is inefficient, given the precision and tuning needed on the developer's part. Hardcoding is also not a generalizable solution, particularly when the task involves interactions with objects or when the task involves different variations in tools. The ability to more generally learn tool use is important when considering the construction of more general-purpose robots that can adapt and learn to accomplish a wider variety of tasks in a shorter period and with limited human assistance.

\begin{figure}[h]
\centering
 \includegraphics[scale=0.45]{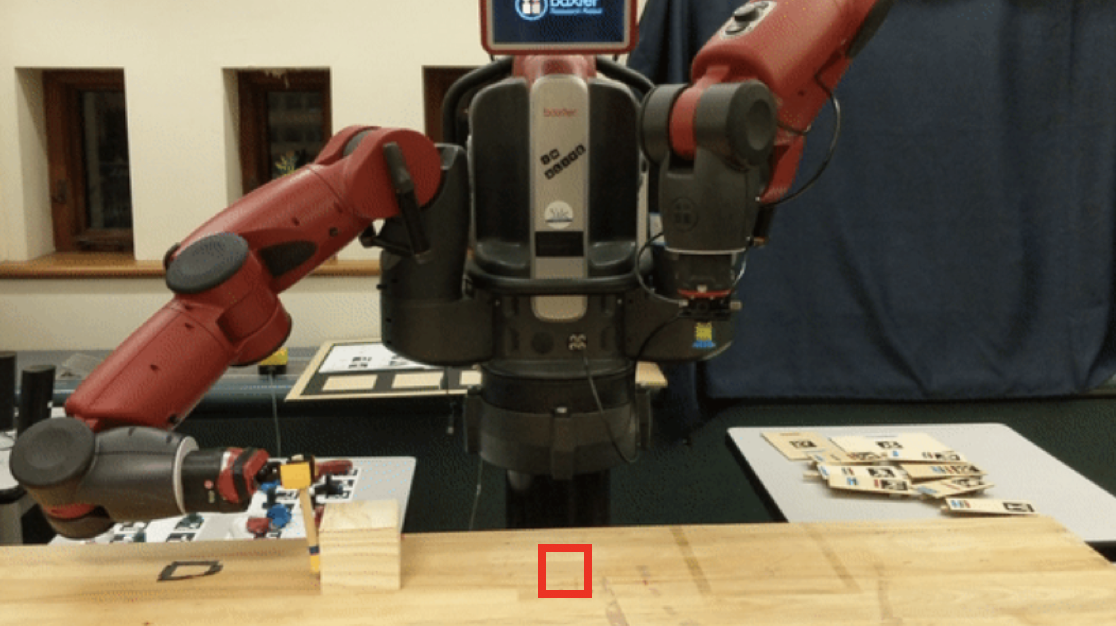}
 \caption{Given a tool and task, there are many solutions to accomplishing the task, which is to push the box to the location corresponding to the red box. We aim to create a setup that would facilitate the learning of more general tasks.}
\end{figure} 

Our primary contributions towards this problem of robot tool use are three-fold: (1) extending an inverse kinematics model with an additional fixed joint for the tool picked up, (2) a novel Baxter simulation model calibrated against the physical system and set up for reinforcement learning, (3) an architecture for using the extended inverse kinematics model with a learned policy such that the overall system can determine an action trajectory that is robust to variable tool length. 

\section{Background}

\subsection{Baxter Robot}
In order to model the usage of human tools appropriately, a robot with a movement capacity and dexterity similar to that of a human is required. The humanoid dual-armed Baxter robot, a product of Rethink Robotics, is designed to mimic human action and ensure compliant operation. It is equipped with two 7-DOF arms, interchangeable gripper mechanisms, and a variety of sensing mechanisms, including several cameras.\cite{aboutBaxter} This platform is ideal for developing a system for tool usage, as it both allows for the manipulation of tools in a human way and ensures that the platform will not damage itself with any tool as a result of its compliance.
\vspace{2mm}
\subsubsection{ROS}
The Baxter robot is controlled with the Robotic Operating System (ROS). ROS is a flexible framework for writing robot software. It is a collection of tools, libraries, and conventions that aim to simplify the task of creating complex and robust robot behavior across a wide variety of robotic platforms.\cite{ros}
\vspace{2mm}
\subsubsection{Inverse Kinematics}
Within the ROS framework, Rethink Robotics provides a control system for Baxter based on angle values for each of the seven joints in each arm. In order to direct the arms to locations in three-dimensional space, an inverse kinematic solver is needed. Creating a system for determining inverse kinematic solutions is a non-trivial task, as there exists an infinity of valid paths from one valid position to another, and not all positions are valid. For the purposes of this project, we used an existing implementation of an inverse kinematic solver\cite{ik} and extended it to allow for tool utilization. The origin of this inverse kinematic system is the center of the robot in the y and z-axes and directly in front of the robot in the x-axis. 

\begin{figure}[h]
\centering
 \includegraphics[scale=0.4]{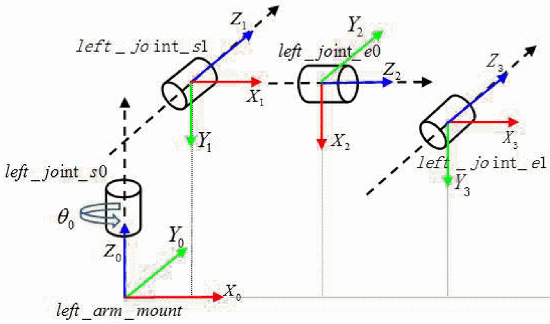}
 \caption{Graphical representation of the complexity of an inverse kinematic solution for the first three joints in the arm. Ju, Zhangfeng, Chenguang Yang, and Hongbin Ma. "Kinematics modeling and experimental verification of Baxter robot." In Control Conference (CCC), 2014 33rd Chinese, pp. 8518-8523. IEEE, 2014.}
 \label{fig:ik}
\end{figure} 

\subsection{Tool Use}
The area of robot tool use has been explored in a number of domains, including computer vision and robotics. Various studies including \cite{b2} have explored how a robot should position its grasp in a task-oriented manner when picking up a tool. In \cite{b2}, the authors propose a self-supervised simulation training approach where tools of various shapes would be procedurally generated. They experimented with two tasks, hammering and sweeping, and their simulation model successfully transferred grasp orientations to the real world. However, action trajectories to accomplish the tasks were hardcoded after the grasp was determined. In \cite{b3} the focus was on tool affordances, that is learning the implications of a tool so tools can be applied to tasks. They propose a set of tool descriptors and investigate the suitability of the tool for various actions through a combination of gathering real-world data with the iCub robot and simulation training. Our work aims to learn a policy that determines an action trajectory for task completion using a tool but is also robust enough to handle variable tool lengths. 

\subsection{Reinforcement Learning for Robot Systems}
Recent approaches for learning robot tools heavily involve deep reinforcement learning in simulation. There are many expected problems with training a model in simulation and transferring it to the physical system. Often, the simulation model is not well calibrated against the real world due to wrong measurements or unfeasibility in determining physical constants like the coefficient of friction. In \cite{b6}, randomizing the simulation environment variables like physical constants, lighting conditions, and camera positions was shown to train a policy that successfully transfers to the real world, which is treated as another randomized environment. In \cite{b5}, a self-supervised approach for learning representations and robotic behaviors from unlabeled videos manages to perform task execution and human imitation. Their method of following human demonstrations in learned representations enabled more efficient reinforcement learning and more effective transfer. Our work aims to learn a policy that successfully transfers to the physical system and is robust to variable tool length while circumventing the problems of training intensity in domain randomization or the extensive data requirements of video imitation. This is accomplished by combining well-established solvers for inverse kinematics with a learned policy to determine a robust and transferrable action trajectory.

\section{Methods}

\subsection{Task Design and Architecture}
Our task is for Baxter to use a variable length tool that it has picked up to push a wooden box across the table to a target location in 3D space. Baxter will use an action trajectory determined by a learned policy for task execution. Our model has three components: (1) extending the existing kinematics solver to account for a new effective joint upon picking up the tool in the real world, including length detection of the tool; (2) learning a policy for successful task execution in simulation; (3) transferring the learned policy to the real world to determine a successful action trajectory for task execution on the physical system.

\begin{figure}[h]
\centering
 \includegraphics[width=0.5\linewidth]{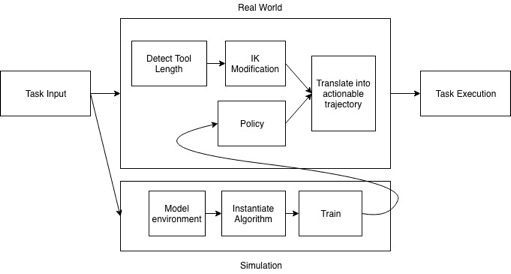}
 \caption{Workflow overview: Given a specific task, the inverse kinematics model is extended on the real robot for a determined tool length, and a policy is learned in simulation. The policy and the modified inverse kinematics model are used to determine an action trajectory that is different depending on the tool length, and the action trajectory is passed to the robot for execution.}
 \label{fig:workflow}
\end{figure} 

The following assumptions were made about the environment when designing the task and associated experiments. We assume that the robot always moves to a fixed location on the table to pick up the tool, the tool is always at that fixed location, and the tool is always grasped at the same orientation. In the simulation, the tool length is a determined constant, but the policy is combined with the detected physical tool length using our extended inverse kinematics model to determine an action trajectory in the real world.

\subsection{Extended Inverse Kinematics Model}
We sought to extend the inverse kinematic model by adding the capacity to move the tip of an arbitrary tool to any valid pose, using the same interface as moving the gripper on the end of the arm. In order to achieve this, we needed two components: a system to detect the length of the tool as gripped, and a system that, given this length, could recalculate a new position for the gripper along an arbitrary axis. By directing the gripper to the offset position, we achieve the functionality of directing the tooltip to an arbitrary position. 

An advantage of this approach is that it allows for any valid path learned in simulation to be directly transferred to the physical robot. Further, since the locations correspond precisely between simulation and reality, training the simulated Baxter to accomplish a task with a tool of a given length will generate a valid. This valid path can then be offset by the difference in the known length of the tool the system is trained on and the measured length of the tool the robot uses in reality. This allows for tool usage of a tool of arbitrary length with no retraining. 

\vspace{2mm}
\subsubsection{Detecting Tool Length}
We utilized basic computer vision techniques in OpenCV to detect the tool length. One key assumption that we made is that the tool was generally long and straight, so we do not account for tools that are more oddly shaped. 
\begin{figure}[h]
\centering
 \includegraphics[scale=0.6]{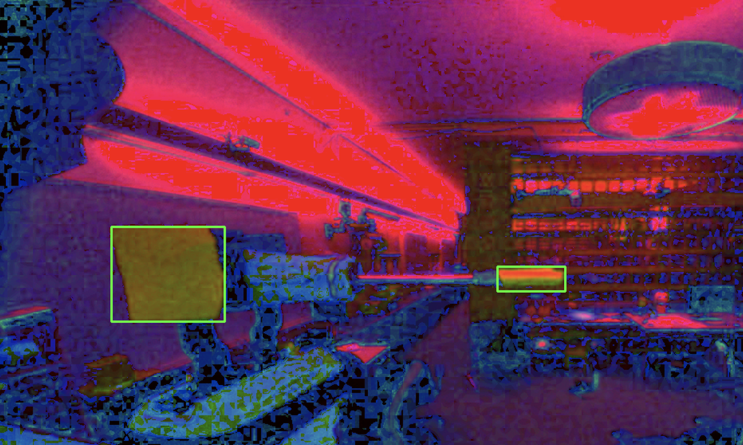}
 \caption{Given three images of the tool from different orientations, we detect its length through computer vision techniques.}
 \label{fig:result2}
\end{figure} 

We programmed Baxter to move to three different orientations: one where the tool is horizontal with the gripped point on the left, another where the tool is horizontal but the gripped point on the right, and another where the tool is vertical with a gripped point on the bottom. Due to the inherent noise in the Baxter system, we took three measurements and then, when extending the kinematics, used the average length found for a more robust number.

The process used to detect the length in each image is as follows: the end of the gripper and the tip of the tool were tagged with a bright orange color that stands out against the background of the room. The image is then converted into HSV space to accentuate the color for detection further. The image is then masked, so anything outside of an orange color is filtered out. We draw bounding rectangular boxes around any clusters of orange in the image and take the two largest boxes, which correspond to the points on the tool because we ensure there are no other large orange-colored objects in the room. Then, we determine the pixel distance between the ends of the two boxes and scale it to fit the measurements in real life.

\vspace{2mm}
\subsubsection{Determining the New Position}
Once the length of the tool has been found, the tip of the tool can be moved to an arbitrary location if the gripper is moved to a location offset by the length of the tool along the correct axis. This can be accomplished by subtracting a three-element vector from the target location, where the appropriate portion of the offset is present in each of the x, y, and z-axis. Determining this vector is trivial for cases in which the offset is entirely present in a single axis, as is the case for the default orientation of the Baxter robot's gripper. In more complex cases, the orientation of the gripper is described using a quaternion. Quaternions are commonly used to describe rotations, and the quaternion associated with the target position can be utilized in this same way. An initial offset vector with the entirety of the offset in the x-axis, $v$ is initialized and rotated by the target quaternion $q$ according to the standard equation
\begin{equation}
    v' = q v q^{-1}
\end{equation}
The new vector $v'$ is then subtracted from the target position, and the gripper is directed to this location. 

\subsection{Training in Simulation}
Training in simulation is essential for our model to learn a policy for determining an action trajectory. We use established software tools for reinforcement learning to set up the simulation model, and we create a novel reinforcement learning environment for the Baxter, simulation model.

In the following subsections, we provide details about the software architecture used to set up the training environment, the environment itself, and the learning algorithms we selected for experimentation.

\vspace{2mm}
\subsubsection{Software Architecture}
We model Baxter in simulation using the MuJoCo physics engine\cite{b7}. Although the favored simulation engine for Baxter is Gazebo, we used MuJoCo instead due to anecdotal evidence of Gazebo being unable to handle object-object interactions and objects slipping out of Baxter's grasp when moving, features that are crucial to our task. Our model of Baxter is based on Rethink Robotics's MuJoCo model but has been modified to match our physical system more closely via calibration. The MuJoCo model has wrapped into an OpenAI Gym\cite{b8} robotics environment for reinforcement learning using the much-py Python interface. We use OpenAI Baselines\cite{b9} for implementations of reinforcement learning algorithms.

Despite significant calibration, it is unavoidable that the simulation is an approximation of the physical system. For example, Baxter's joint angles are more constrained on the physical system than on simulation due to stricter collision avoidance thresholds. Furthermore, physical coefficients like friction are difficult to measure and are likely to change over time. Despite the expected reality gap, it remains important that we train in simulation. Deep reinforcement learning requires training data on the order of millions, and training on the physical system is unfeasible. In \cite{b4}, robot-grasping on the physical system is learned after 50K iterations over 700 hours. To address this issue, we only transfer the policy using the learned trajectory, defined by a sequence of coordinates in 3D space and quaternions. 

To the best of our knowledge, our setup is the first to utilize such an architecture involving MuJoCo and Gym to train the Baxter robot in simulation. 

\vspace{2mm}
\subsubsection{Environment}
Within the MuJoCo framework, the model of the robot is specified in an XML file obtained from Rethink Robotics\cite{b10}. This facilitates the incorporation of the robot model into our simulation, along with the required objects for our environment, like the table, box, and tool. In addition, we weld MuJoCo's motion capture (mocap) object to the right gripper of the Baxter model and enforce an equality constraint to the mocap object. This allows us to specify coordinates in 3D space for the mocap object to which the Baxter model in a simulation will force its gripper position using the in-built inverse kinematics solver. As a result, each action of the gripper can be specified as a change in the mocap position.

Within the Gym framework, we create an environment for the Baxter system modeled after the example Fetch environment in the Gym. Fetch is a one-armed robot whose task is to use its end effector to move a puck to a target location. The Markov Decision Process for the Baxter agent, adapted from Fetch to the Baxter environment, is the following: at each time step $t$, the agent observes the current state $s_t$, which includes the:
\vspace{1mm}
\begin{itemize}
 \item Goal box position $x*$
 \item Current box $x$
 \item Tool position $y$
 \item Gripper position
 \item Gripper orientation as a quaternion
 \item Joint angles for each actuator
 \item Joint velocities for each actuator
\end{itemize}
\vspace{1mm}
The agent samples an action $a_t \sim \pi_\theta(\cdot|s_t)$, which determines the change in position of the mocap object in 3D space and the change in quaternion for the gripper. This action space is thus a continuous action space in $\mathbb{R}^7$. The agent receives a reward $r_t = - (d(x,x*) + d(x, y))$ where $d$ is the Euclidean distance. This agent is then trained with various reinforcement learning algorithms. The goal condition is defined as $d(x,x*) + d(x,y) \leq \alpha$ where $\alpha = 0.05$m is a specified threshold. Note that $x*$ and $x_0$ are different only along the y-dimension, so the task is simply to push a box horizontally across the agent's body. Each episode runs until the goal condition is achieved or 100 timesteps are run. We refer to this environment as Env1.

\vspace{2mm}
\subsubsection{Reinforcement Learning Algorithms}
To attain the best policy, we train the agent on various reinforcement learning algorithms. 

We employ model-free policy gradient methods that perform gradient ascent to maximize the performance of the agent. The algorithms are A2C (Advantage Actor-Critic)\cite{a2c}, TRPO (Trust Region Policy Optimization)\cite{trpo}, and PPO (Proximal Policy Optimization)\cite{ppo}. These are on-policy methods, so each update only uses data collected from the most recent policy, and no hindsight or experience is utilized. We also employ DDPG, a model-free algorithm that combines policy gradient and Q learning that learns on-policy and off-policy concurrently and uses each to improve the other. We choose the best model for transfer to the real world.

\textbf{A2C} is the synchronous implementation of A3C (Asynchronous Advantage Actor-Critic). A3C utilizes a global network with multiple worker agents each with their network parameters, allowing for training that is more independent and diverse. The actor-critic model aims to combine the benefits of value-iteration and policy-iteration methods. Given the on-policy value function 

$$V^{\pi}(s) = E_{\tau \sim \pi}[R(\tau)|s_0 = s]$$
\noindent
and the on-policy action-value function

$$Q^{\pi}(s,a) = E_{\tau \sim \pi}[R(\tau)|s_0 = s, a_0 = a]$$
\noindent
the advantage function

$$A^{\pi}(s,a) = Q^{\pi}(s,a) - V^{\pi}(s)$$
\noindent
quantifies the relative advantage of the action $a$ in state $s$ over randomly selecting an action according to $\pi(\cdot|s)$ and then acting according to $\pi$ thereafter. A3C utilizes asynchronous updates of the global network parameters, whereas A2C uses a synchronous update. 

\textbf{TRPO} employs the actor-critic paradigm but with a differing policy update

$$\begin{array} { r } { \theta _ { k + 1 } = \arg \max _ { \theta } \mathcal { L } \left( \theta _ { k } , \theta \right) } \\ { \text { s.t. } \overline { D } _ { K L } \left( \theta | | \theta _ { k } \right) \leq \delta } \end{array}$$
\noindent
where $\mathcal { L } \left( \theta _ { k } , \theta \right)$ is the surrogate advantage measuring how well policy $\pi_\theta$ performs relative to the old policy $\pi_{\theta_{k}}$:

$$\mathcal { L } \left( \theta _ { k } , \theta \right) = \underset { s , a \sim \pi _ { \theta _ { k } } } { E } \left[ \frac { \pi _ { \theta } ( a | s ) } { \pi _ { \theta _ { k } } ( a | s ) } A ^ { \pi _ { \theta _ { k } } } ( s , a ) \right]$$
\noindent
and $D _ { K L } ( \theta \| \theta _ { k } )$ is the average KL divergence between policies for states visited by the old policy.

$$D _ { K L } ( \theta \| \theta _ { k } ) = \underset { s \sim \pi \theta _ { k } } { \mathrm { E } } \left[ D _ { K L } \left( \pi _ { \theta } ( \cdot | s ) \| \pi _ { \theta _ { k } } ( \cdot | s ) \right) \right]$$

\textbf{PPO} employs a similar to TRPO, except the policy update incorporates the KL-divergence constraint directly into the advantage function. The policy update is 

\begin{equation*}
 \theta _ { k + 1 } = \arg \max _ { \theta } \operatorname { E } _ { s , a \sim \pi _ { \theta _ { k } } } \left[ L \left( s , a , \theta _ { k } , \theta \right) \right]
\end{equation*}
\noindent
and $L$ is given by

\begin{align}
    L(s, a, \theta_k, \theta) &= \min \left( 
    \frac{\pi_\theta(a|s)}{\pi_{\theta_k}(a|s)} A^{\pi_{\theta_k}}(s, a), \right. \notag \\
    & \quad \left. \operatorname{clip} \left( 
    \frac{\pi_\theta(a|s)}{\pi_{\theta_k}(a|s)}, 1 - \epsilon, 1 + \epsilon 
    \right) A^{\pi_{\theta_k}}(s, a) 
    \right) 
\end{align}

\textbf{DDPG} conducts off-policy Q-learning by minimizing the following loss function using stochastic gradient descent:

\begin{align}
L(\phi, \mathcal{D}) &= \mathbb{E}_{(s, a, r, s', d) \sim \mathcal{D}} \left[ \left( Q_{\phi}(s, a) - \right. \notag \right. \\
& \quad \left. \left( r + \gamma (1 - d) Q_{\phi_{\text{targ}}}(s', \mu_{\theta_{\text{targ}}}(s')) \right) \right)^{2} \notag \\
&= \mathbb{E}_{(s, a, r, s', d) \sim \mathcal{D}} \left[ \left( Q_{\phi}(s, a) - \hat{Q} \right)^{2} \right] \label{eq:loss_function}
\end{align}

\noindent
with $\mu_{\theta_targ}$ as the target policy. DDPG also conducts on policy policy updates by performing gradient ascent on

$$\max _ { \theta } \underset { s \sim \mathcal { D } } { \mathrm { E } } \left[ Q _ { \phi } \left( s , \mu _ { \theta } ( s ) \right) \right]$$

\subsection{Simulation to Real Transfer}
To determine an action trajectory from the learned policy after a training session, we run the model for 100 complete episodes and record the gripper position in 3D space and gripper orientation specified by the quaternion at each timestep, taking the average positions over the 100 complete episodes. Each gripper coordinates, and quaternion is then fed into our extended inverse kinematics model by subtracting from them an offset specified by the determined tool length in the orientation specified by the quaternion. The resulting gripper coordinate and original quaternion are then used as positions and orientations for which the Baxter physical system has to reach. We use a constant wait time between time steps to allow the robot to solve for the next point and move into position. 

\section{Experiments}

We set up experiments for testing (1) our extended inverse kinematics model, (2) training in simulation, and (3) simulation to real transfer.

\subsection{Extended Inverse Kinematics Model}
Once implemented, we confirmed the accuracy of the extended inverse kinematic model by repeatedly directing the solver to move the gripper with no tool, the gripper with a shorter tool, and the gripper with a longer tool to a selection of positions. We measured the difference between the requested position and the position reached and tabulated the error and standard deviation. These results are presented in Figure \ref{fig:ikextension}. 

\subsection{Training in Simulation}

\vspace{2mm}
\subsubsection{Environments}
We trained models on the Env1 environment for 1,250,000 timesteps. Training each model takes about seven hours on 32 Intel(R) Xeon(R) CPU E5-2620 v4 @2.10 GHz CPUs. We also trained models on a modified environment Env2, which is exactly the same as Env1 but with reward function $r_t = -d(x,x*)$. We eventually ignored models trained on Env2 as the best model in Env2 learned to push the box using the gripper instead of the tool, and other models did not push the box at all. The training results are presented in Figure \ref{fig:algos}.

\vspace{2mm}
\subsubsection{Fine-Tuning}
We also defined a modified environment, Env3, which is exactly the same as Env1 but with the new goal box position specified as $x + 0.10$m. We fine-tuned the model trained on Env1 on Env3 for a further 225,000 timesteps. We also compared this fine-tuned model to a model trained on Env3 from scratch for 1,475,000 timesteps. The training results are presented in Figure \ref{fig: finetune}. We investigated this only for the model trained using PPO.

\section{Results}

\subsection{Extended Inverse Kinematics Model}
\begin{figure}[h]
\centering
 \includegraphics[scale=0.2]{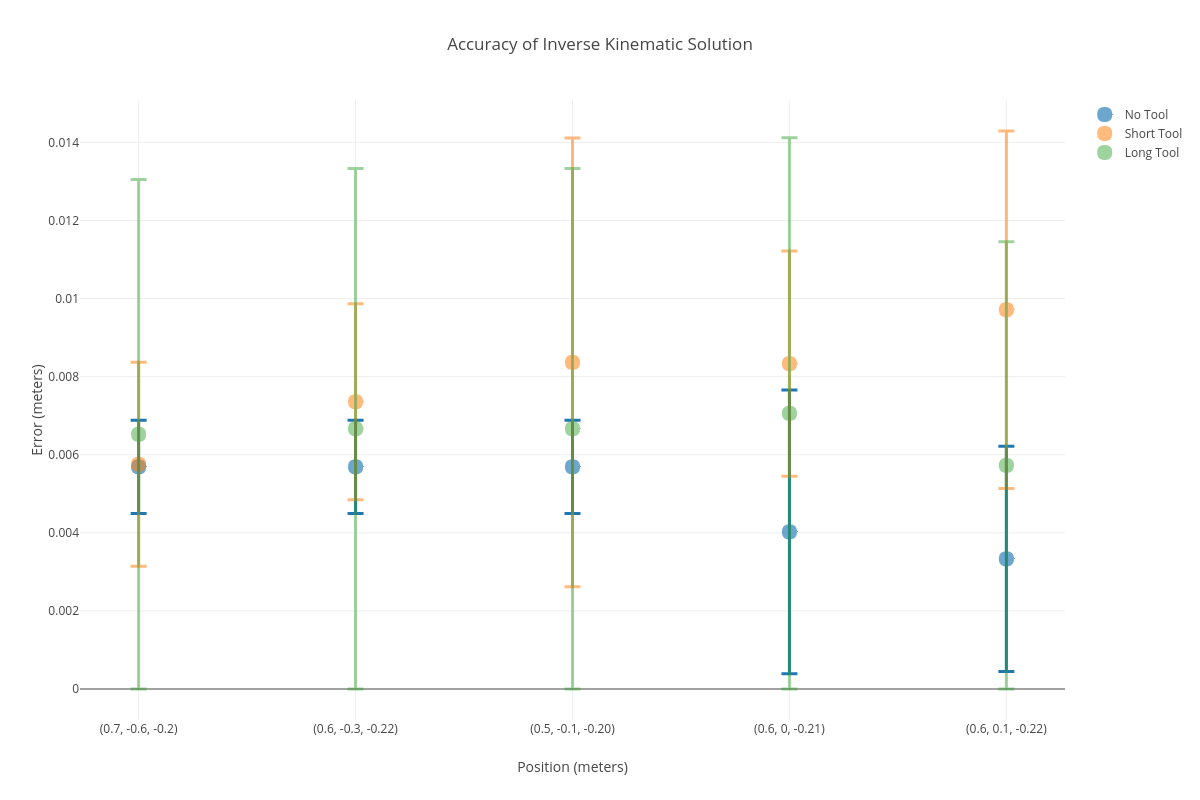}
 \caption{Plot of the difference between the target position and the position reached by Baxter's inverse kinematics solver with no tool, a long tool, and a short tool in meters. The accuracy decreases when the solver is extended with a tool but remains under a centimeter.}
 \label{fig:ikextension}
\end{figure}

The extended inverse kinematic solver performed relatively well, with an error of less than 1cm. Across each of the target locations, neither tool was, on average, greater than one centimeter away from its goal. While the performance of the extended model was worse than the gripper alone, this measurement does not account for the error in detected tool length from the computer vision system. We expect the performance of the inverse kinematic extension to increase significantly with a more sophisticated image processing system. These results can be seen in Figure \ref{fig:ikextension}. 

\subsection{Training in Simulation}

\vspace{2mm}
\subsubsection{Comparison Across Algorithms}
\begin{figure}[h]
\centering
 \includegraphics[scale=0.55]{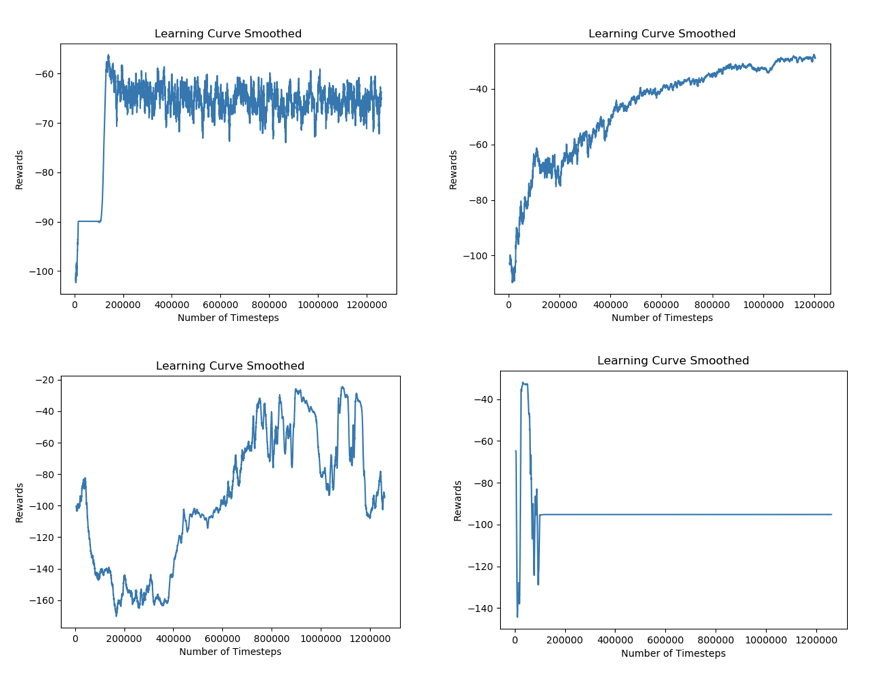}
 \caption{Reward plots for the four algorithms: top left A2C, top right TRPO, bottom left PPO, and bottom right DDPG.}
 \label{fig:algos}
\end{figure}

We plot the reward as a function of time steps for each algorithm during the training process in Figure \ref{fig:algos}. Furthermore, we report the average distances of the box position at the end of an episode to the goal position over 100 episodes for each model. A2C and DDPG exhibit the fastest convergence behavior. However, A2C exhibits unstable movements and rotations and attains a 40.9cm average distance from the goal position. In addition, DDPG results in a policy where the agent simply moves its gripper away from the box. The box remains at its original position, and thus, the DDPG model attains a 35.1cm average distance from the goal position. The unusual training behavior of DDPG is likely due to insufficient hyperparameter optimization. 

TRPO exhibits a smooth and expected reward plot and successfully learns a policy that moves the box across to the desired goal position. However, it does so with an unreplicable swap in orientation halfway in motion and is thus not suitable for transfer. PPO results in the best policy that successfully pushes the box towards the goal position in a stable manner. The TRPO model attains an 11.9cm average distance, and the PPO model attains a 7.74cm average distance from the goal position. Both models push the box horizontally for a travel distance of 23.1cm and 27.26cm, respectively.

\vspace{2mm}
\subsubsection{Action Trajectories}
We plot the generated trajectories for the models trained on Env1 in Figure \ref{fig: trajectory}. The A2C model exhibits no clear trajectory, even when taking their average. The DDPG model exhibits a clear trajectory, but as discussed above, the trajectory moves in a direction away from the box. The TRPO model exhibits a clear trajectory with a parabolic arc, and similarly so for the PPO model. Note that since the task is only to push the box horizontally across the y-dimension, we would expect to see a flat trajectory. However, both TRPO and PPO models exhibit trajectories that go up near the end of the episode, and this corresponds to our observations when the simulation is rendered.

\vspace{2mm}
\subsubsection{Comparison for Fine-Tuning}
We only run the fine-tuning for the PPO model, which is our best model. The results are shown in Figure \ref{fig: finetune}. We observe that the fine-tuned model exhibits a higher reward than the from-scratch model and performs significantly better in qualitative terms. The fine-tuned model is able to push the box further than the original model and also in a stable manner. The from-scratch model exhibits stable behavior but does not push the box.

\begin{figure}[h]
\centering
 \includegraphics[scale=0.4]{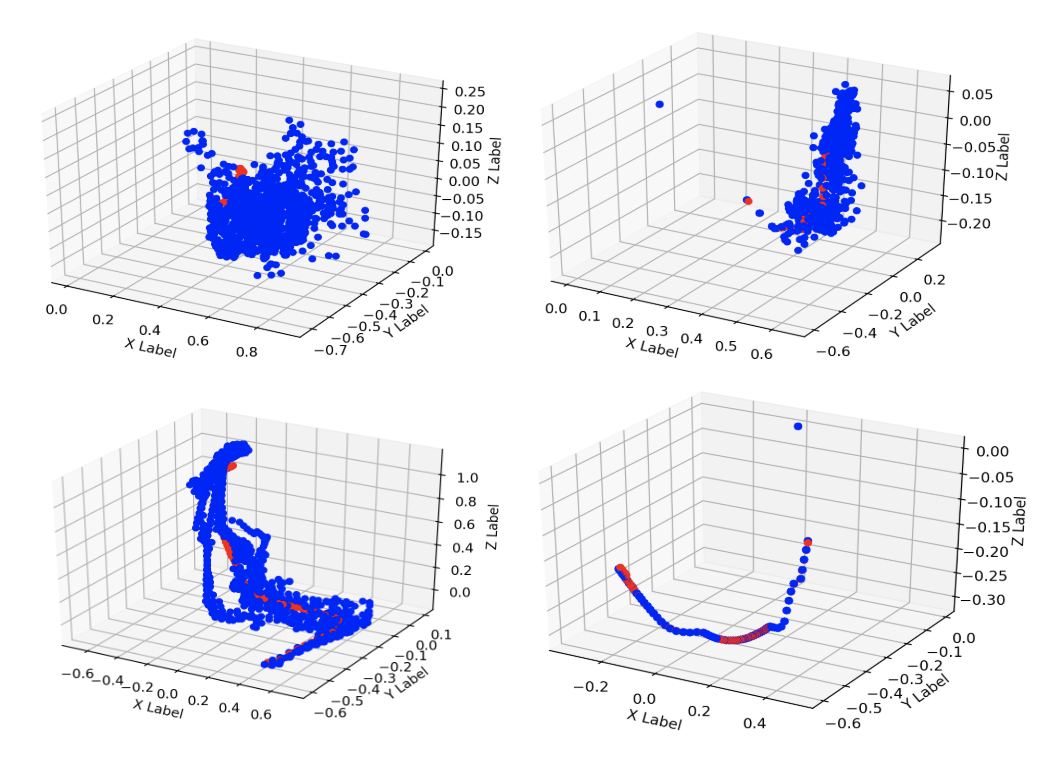}
 \caption{3D plots of 10 generated trajectories from the four algorithms used for transferring to the real robot with blue dots corresponding to generated points and red dots corresponding to an averaged point of the ten runs. Top left A2C, top right TRPO, bottom left PPO, and bottom right is DDPG.}
 \label{fig:trajectory}
\end{figure} 

\begin{figure}[h]
\centering
 \includegraphics[scale=0.6]{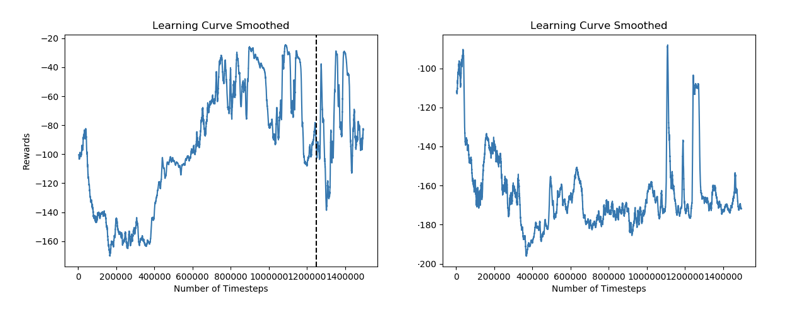}
 \caption{Reward plot comparison between fine-tuned results for PPO compared to training on Env3 from scratch. The dotted line indicates where training on Env1 stopped and fine-tuning began on Env3.}
 \label{fig:finetune}
\end{figure} 

\subsection{Simulation to Real Transfer}
We found a mixed bag of simulation to real-world integration results when we ported the learned trajectories from simulation to be run on the physical system. Below we present general qualitative observations about the generated action trajectories from simulation, which are comprised of points in 3D space and quaternions when we attempted to execute them on the physical system.

\begin{table}[h]
    \centering
    \caption{Observations for Different Algorithms}
    \begin{tabular}{@{}c p{7cm}@{}}  
        \toprule
        Algorithm & Observations \\ \midrule
        A2C & The learned points immediately led to failure when executing the positions and quaternion orientations. \\
        TRPO & Successfully executed the learned actions, but due to the twisted orientation of the pose, it was not successful in accomplishing the goal. \\
        PPO & This was the most successful algorithm that could be transferred from simulation to real life by running through all points. \\
        DDPG & Was able to execute half of the points in the learned trajectory but was not successful in accomplishing the task. \\ 
        \bottomrule
    \end{tabular}
    \label{tab:algorithm-observations}
\end{table}

Aligning with the training results and observations from previous sections, PPO proved to be the transferable algorithm. The other three algorithms were all unable to complete the task on which it was trained. This is not surprising because A2C and DDPG were not successful in simulation, and TRPO exhibited unreplicable re-orienting behavior. This was not improved by taking the average action trajectory over 100 episodes. We present the results of 10 runs of the PPO algorithm with measured box travel distance from the starting point and will compare to the actual distance in simulation, done for two tools of different lengths.

\begin{figure}[h]
\centering
 \includegraphics[scale=0.37]{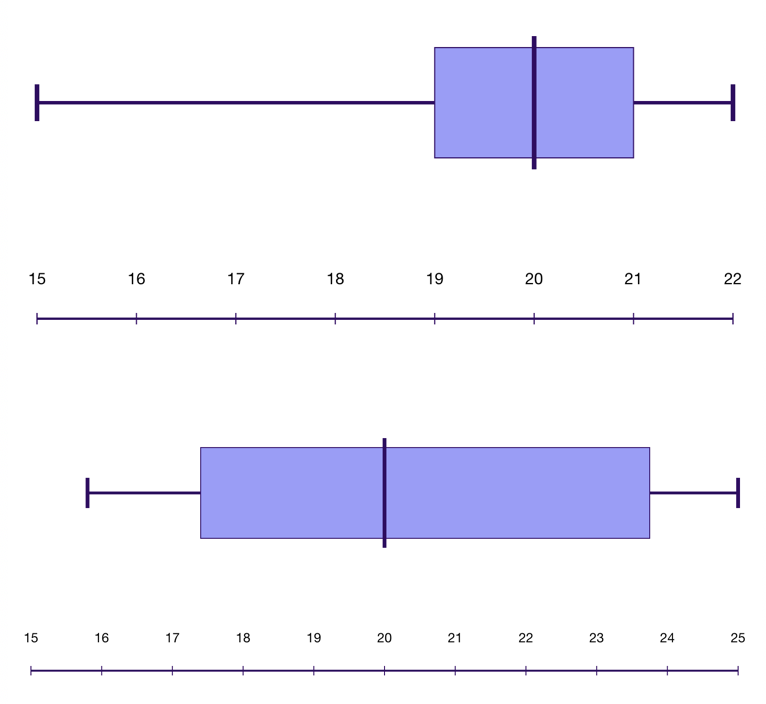}
 \caption{Box plot of the distance box distance moved in cm over ten runs of the same PPO smoothed trajectory with two different tool lengths. The target simulation distance to attain is 25 cm. The top plot is the result of a longer 17.5cm tool, and the bottom plot is the result of a shorter 12.5cm tool.}
 \label{fig:boxplot}
\end{figure} 

In Figure \ref{fig:boxplot}, we provide numerical results of the simulation to real transfer over ten runs on two different tools. We observe that the model moves the box an average of 19.9cm for the longer 17.5cm tool and an average of 19.7cm for the shorter 12.5cm tool. For comparison, the average simulation travel distance is 27.26cm. Note that this average was taken over 100 episodes. This gap in simulation and transfer is expected. However, our model performs virtually identically despite different tool lengths, showing that the learned policy is robust to variable tool lengths in determining equivalent action trajectories.

We find that there is significant noise in the runs despite providing Baxter with the same set of trajectory each time. One would not ordinarily expect this to occur, as the location correspondence between simulation and reality is nearly if not exactly 1:1. In the following discussion, we provide some possible explanations as to where the differences in results between the real world and simulation might arise.  

\section{Discussion}
From the results presented above, we found that the model we developed performed reasonably well. The extended inverse kinematic solver proved to be a robust solution for moving a tooltip of an arbitrary tool to an arbitrary point in three-dimensional space, and it proved able to precisely replicate the valid positions of a learned solution to a given task with a fixed tool in simulation with an arbitrary tool in reality.  There were, however, some limitations encountered.  

\subsection{Limitations}
The main limitations of our approach are found in simulation results, specifically in transferring the learned paths from the simulator to the real robot. There are two main hurdles to this task. The first is that in simulation, the modeled tool is bound directly to the gripper. This means that any pliancy or tendency for the tool to slip is not captured in the simulation. Further, this means that even though we account for the different lengths of the tool in simulation and the tool as held by the robot, the tool may be held in a slightly different location laterally. The gripping surface of the gripper provides several centimeters of sufficient grip locations, and if the tool were held at either of the extremes, this would explain much of the variability in path length. 

The second hurdle also relates to simulator physics. In the simulator, Baxter is allowed to move to any reachable position according to the published range of motion of each joint. However, this system does not perfectly account for collisions of the robot arm with itself and with other objects. Moreover, the inverse kinematic solver is somewhat more conservative about reachable positions than the exhaustive simulation. As a result, many of the points and orientations generated by the simulation are not reachable by the inverse kinematics solver. Attempting to reach these positions can cause the arm to lock up and fail to reach later valid states. 

In order to overcome these limitations, we suggest that modifications be made to the simulator to reflect the reality of the robot's operation better. Using the results of the current simulation is also possible if the generated path is smoothed by sampling out of the full list of simulation locations at each time step to create a simpler path, and at the same time, only valid positions, in reality, are selected for this path. This can be accomplished with a relatively simple filtering scheme. 

Detection of object length is also currently limited to objects whose entire length fits within the view of the Baxter robot's head camera for the three positions we selected. This restriction can be overcome by moving the positions further away from the head or implementing a more sophisticated image processing system capable of stitching together multiple frames to accommodate tools of greater length. 

\subsection{Future Work}
The methods and results that we have presented in this paper lend themselves to a wide array of extensions for future work toward the goal of a more general tool-learning approach. Since we only learned the sweeping task using one shape and type of tool, it would be interesting to model and learn about other types of tools as well. The idea would be to have certain tasks that will perform better with certain tasks and for the robot to be able to decide which tool is best for the task.

Additionally, with regards to the simulation and real transfer, it would be interesting to quantify the acceptable movement ranges of the real robot and only allow the simulation to learn given those allowable values of position and orientation. This would facilitate the transfer of information. 

\section{Conclusion}
We developed a pipeline for which a robot can detect tool length and extend its inverse kinematics model to account for the tool, learn to accomplish a known task using a tool through training in simulation, and when given a similarly shaped tool, the action trajectory can be determined to fit the new tool length from an extension of the inverse kinematics model. We successfully trained in simulation on four different algorithms to determine the algorithm that could provide the best transferable action trajectory. We observed expected gaps in simulation to real transfer but also observed that the learned policy, when combined with the extended inverse kinematics model, is able to determine an action trajectory that is robust to variable tool lengths. We acknowledge certain limitations in our simulation setup and the contribution of noise in the real robot, which led to larger errors in the distance the box moved in comparison to the simulated motion. However, with some modifications, we believe this setup is a viable path for further exploration into more general robot tool learning.


\begin{thebibliography}{00}
\bibitem{ros}"About ROS. " http://www.ros.org/about-ros/.
\bibitem{b2} Fang, Kuan, Yuke Zhu, Animesh Garg, Andrey Kurenkov, Viraj Mehta, Li Fei-Fei, and Silvio Savarese. Learning task-oriented grasping for tool manipulation from simulated self-supervision. arXiv preprint arXiv:1806.09266 (2018).
\bibitem{aboutBaxter} Ju, Zhangfeng, Chenguang Yang, and Hongbin Ma. "Kinematics modeling and experimental verification of baxter robot." In Control Conference (CCC), 2014 33rd Chinese, pp. 8518-8523. IEEE, 2014.
\bibitem{b3} Mar, Tanis, Vadim Tikhanoff, and Lorenzo Natale. What Can I Do With This Tool? Self-Supervised Learning of Tool Affordances From Their 3-D Geometry. IEEE Transactions on Cognitive and Developmental Systems 10, no. 3 (2018): 595-610.
\bibitem{a2c} Mnih, Volodymyr, Adria Puigdomenech Badia, Mehdi Mirza, Alex Graves, Timothy Lillicrap, Tim Harley, David Silver, and Koray Kavukcuoglu. "Asynchronous methods for deep reinforcement learning." In International conference on machine learning, pp. 1928-1937. 2016.
\bibitem{ddpg} Lillicrap, Timothy P., Jonathan J. Hunt, Alexander Pritzel, Nicolas Heess, Tom Erez, Yuval Tassa, David Silver, and Daan Wierstra. "Continuous control with deep reinforcement learning." arXiv preprint arXiv:1509.02971 (2015).
\bibitem{b4} Pinto, Lerrel, and Abhinav Gupta. Supersizing self-supervision: Learning to grasp from 50k tries and 700 robot hours. In Robotics and Automation (ICRA), 2016 IEEE International Conference on, pp. 3406-3413. IEEE, 2016.
\bibitem{ik}Roncone, Alessandro, Olivier Mangin, and Brian Scassellati. "Transparent role assignment and task allocation in human robot collaboration." In Robotics and Automation (ICRA), 2017 IEEE International Conference on, pp. 1014-1021. IEEE, 2017.
\bibitem{b1} Solly Brown and Claude Sammut. Tool use and learning in robots. In Encyclopedia of the Sciences of Learning, pages 3327–3330. Springer, 2012.
\bibitem{ppo} Schulman, John, Filip Wolski, Prafulla Dhariwal, Alec Radford, and Oleg Klimov. "Proximal policy optimization algorithms." arXiv preprint arXiv:1707.06347 (2017).
\bibitem{trpo} Schulman, John, Sergey Levine, Pieter Abbeel, Michael Jordan, and Philipp Moritz. "Trust region policy optimization." In International Conference on Machine Learning, pp. 1889-1897. 2015.
\bibitem{b5} Sermanet, Pierre, Corey Lynch, Yevgen Chebotar, Jasmine Hsu, Eric Jang, Stefan Schaal, Sergey Levine, and Google Brain. "Time-contrastive networks: Self-supervised learning from video." In 2018 IEEE International Conference on Robotics and Automation (ICRA), pp. 1134-1141. IEEE, 2018.
\bibitem{b6} Tobin, Josh, Rachel Fong, Alex Ray, Jonas Schneider, Wojciech Zaremba, and Pieter Abbeel. "Domain randomization for transferring deep neural networks from simulation to the real world." In Intelligent Robots and Systems (IROS), 2017 IEEE/RSJ International Conference on, pp. 23-30. IEEE, 2017.
\bibitem{b7} Todorov, Emanuel, Tom Erez, and Yuval Tassa. "Mujoco: A physics engine for model-based control." In Intelligent Robots and Systems (IROS), 2012 IEEE/RSJ International Conference on, pp. 5026-5033. IEEE, 2012.
\bibitem{b8} Brockman, Greg, Vicki Cheung, Ludwig Pettersson, Jonas Schneider, John Schulman, Jie Tang, and Wojciech Zaremba. "Openai gym." arXiv preprint arXiv:1606.01540 (2016).
\bibitem{b9} Dhariwal, Prafulla, Christopher Hesse, Oleg Klimov, Alex Nichol, Matthias Plappert, Alec Radford, John Schulman, Szymon Sidor, and Yuhuai Wu. "Openai baselines." GitHub, GitHub repository (2017).
Harvard	
\bibitem{b10} "Baxter 1.31." Universal Robots UR5 Robotiq S Model 3 Finger Gripper | MuJoCo Forum. Accessed December 07, 2018. http://www.mujoco.org/forum/index.php?resources/baxter.17/.

\end{thebibliography}
\end{document}